\documentclass{article}
\usepackage{spconf,amsmath,graphicx}
\usepackage{multicol,multirow}
\usepackage{color}

\title{Knowledge-augmented Frame Semantic Parsing with \\ Hybrid Prompt-tuning}
\name{Rui Zhang$^\dag$, Yajing Sun$^\dag$ \thanks{$^\dag$ Contributed equally to this work.}, Jingyuan Yang, Wei Peng$^*$ \thanks{$^*$ Corresponding author.}}
\address{Artificial Intelligence Application Research Center, Huawei Technologies}
\begin{document}
%
\maketitle
\begin{abstract}
Frame semantics-based approaches have been widely used in semantic parsing tasks and have become mainstream. It remains challenging to disambiguate frame representations evoked by target lexical units under different contexts. Pre-trained Language Models (PLMs) have been used in semantic parsing and significantly improve the accuracy of neural parsers. However, the PLMs-based approaches tend to favor collocated patterns presented in the training data, leading to inaccurate outcomes. The intuition here is to design a mechanism to optimally use knowledge captured in semantic frames in conjunction with PLMs to disambiguate frames. We propose a novel Knowledge-Augmented Frame Semantic Parsing Architecture (KAF-SPA) to enhance semantic representation by incorporating accurate frame knowledge into PLMs during frame semantic parsing. Specifically, a Memory-based Knowledge Extraction Module (MKEM) is devised to select accurate frame knowledge and construct the continuous templates in the high dimensional vector space.
Moreover, we design a Task-oriented Knowledge Probing Module (TKPM) using hybrid prompts (in terms of continuous and discrete prompts) to incorporate the selected knowledge into the PLMs and adapt PLMs to the tasks of frame and argument identification. Experimental results on two public FrameNet datasets demonstrate that our method significantly outperforms strong baselines  (by more than +3$\%$ in F1),  achieving state-of-art results on the current benchmark. Ablation studies verify the effectiveness of KAF-SPA. 
\end{abstract}

\section{Introduction}
\label{sec:intro}
\begin{figure}[htbp]
    \centering
    \includegraphics[width=0.95\linewidth]{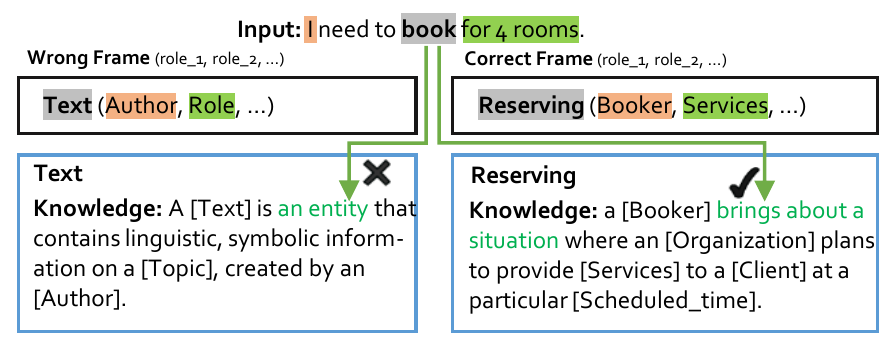}
    \caption{An illustration of the knowledge-augmented frame semantic parsing for a target word ``book". The black boxes represent the predicted frame and related arguments, and the blue boxes underneath represent the frame definitions. Proper frame definitions are beneficial for frame semantic parsing for given inputs. } 
    \label{fig:intro}
\end{figure}

Frame semantics defines the conceptual structure based on FrameNet \cite{baker1998berkeley,das2014frame}, capturing the background knowledge necessary for an agent to understand a situation. 
It is fundamental for many NLP applications such as dialogue systems \cite{chen2013unsupervised}, machine reading comprehension \cite{guo2020frame}, and question answering \cite{shen2007using}. 
The frame semantic parsing task aims at identifying semantic structure evoked in text, consisting of semantic frames and their associated arguments. It remains challenging to disambiguate frame representations triggered by the targeted lexical units in various aspects of contexts.

Recent advances in semantic parsing can be attributed to Pre-trained Language Models (PLMs), which significantly improved the accuracy of neural parsers \cite{kalyanpur2020open, petroni2019language, li2021prefix}. 
Researches treat Frame-Semantic Parsing task as a classification \cite{das2014frame, kshirsagar2015frame, swayamdipta2017frame, bastianelli2020encoding, marcheggiani2020graph} or a generation problem \cite{kalyanpur2020open}. Specifically, \cite{kalyanpur2020open} showed that a pure generative encoder-decoder architecture handily beats the previous state-of-the-art in a FrameNet 1.7 parsing task.
However, these methods treat frames and arguments as discrete labels represented using one-hot embeddings, neglecting the natural bondage among frame elements that give meanings to utterances. 
In fact, FrameNet \cite{baker1998berkeley}, as a lexical database of concepts with annotated meanings based on semantic frames, defines events, relations, entities, and the relevant participants. Researches that learned from an integrated semantic space encompassing frame definition, frame elements, and frame-frame relationships can achieve better frame semantic parsing results \cite{jiang2021exploiting,su2021knowledge,zheng2022double}. 

Despite some progress achieved along this line, the fundamental issue of frame representation disambiguation remains unsolved.  
\begin{figure*}
    \centering
    \includegraphics[width=0.91\textwidth]{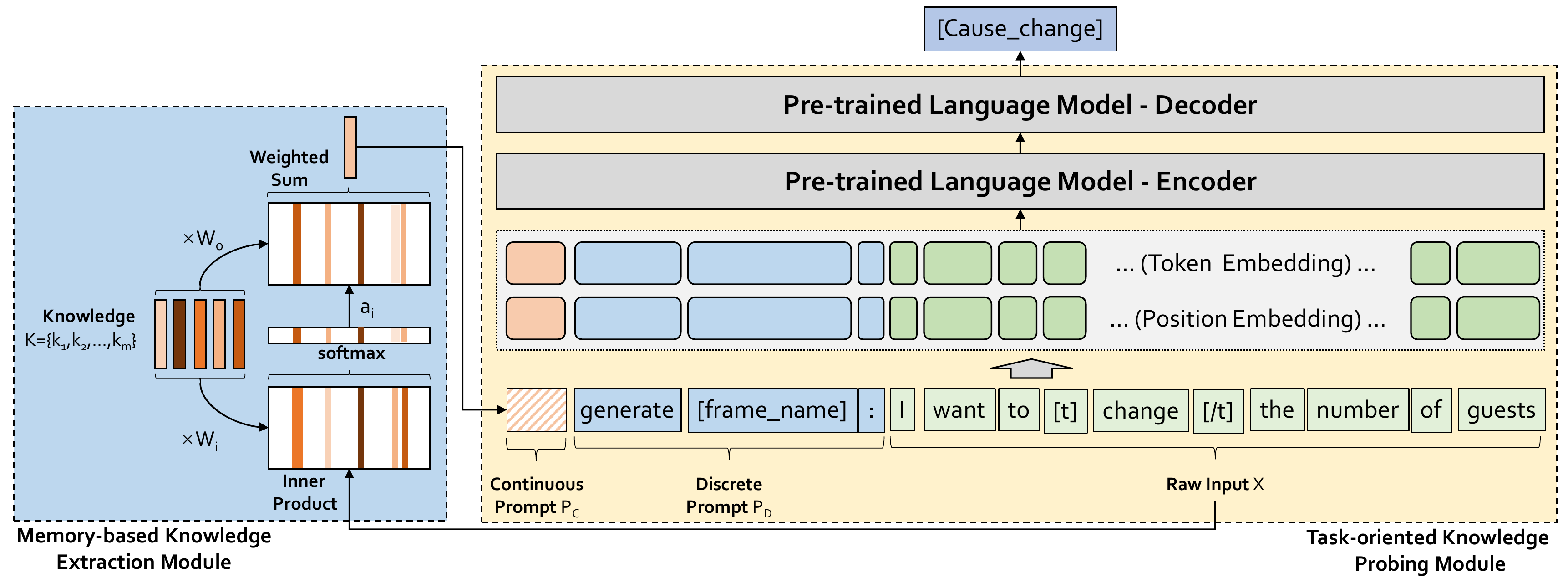}
    \caption{The overall architecture of the proposed KAF-SPA. KAF-SPA consists of two modules: the Memory-based Knowledge Extraction Module (MKEM) and the Task-oriented Knowledge Probing Module (TKPM).} 
    \label{fig:overall}
\end{figure*}
As shown in Figure \ref{fig:intro}, the target word ``book" is linked to two ambiguous frames: \textbf{Text} and \textbf{Reserving}. A PLMs-based approach favors the incorrect frame ``Text" based on our observations. A potential reason is that frequent collocations of ``book'' and related contexts depicting a ``book'' (as a literature of recording information) from the training data may account for the ill-classified frame label (``Text").   The intuition here is to design a mechanism leveraging additional information captured in the semantic frames of FrameNet to alleviate the issue. For example, secondary knowledge supporting an accurate selection of each frame (fonts appeared in the frame definition box in Figure \ref{fig:intro}) can be used by PLMs to create a desirable result during frame semantic parsing.       

This paper proposes Knowledge-Augmented Frame Semantic Parsing Architecture (KAF-SPA) to enhance semantic representation and disambiguate semantic frames. 
Specifically, we design a Memory-based Knowledge Extraction Module (MKEM) to select the most relevant frame definitions and inject the latent knowledge into the PLMs with a continuous learnable knowledge vector. Furthermore, the Task-oriented Knowledge Probing Module (TKPM) is used to incorporate the selected knowledge into the PLMs and adapt PLMs to different subtasks.  Concretely, the knowledge-aware continuous prompt exploits the selected label semantics (from MKEM) to learn the semantic representation of the sentence. Meanwhile, the task-aware discrete prompt is conducted to adapt the PLMs to different frame semantic parsing subtasks.
It is worth noting that we are the first to design hybrid prompts to the adaptation of PLMs in frame semantic parsing.
Finally, We conduct experiments on the FrameNet1.5 and FrameNet1.7 datasets with frame identification and arguments identification tasks. Experiment results demonstrate that our model achieves state-of-art in most metrics. The ablation studies also verify the effectiveness of the method.

\section{Task Definition}

Frame semantic parsing consists of \textbf{frame identification} and \textbf{argument identification} task. 
Given a sentence $X = \{w_0, w_1, \cdots, w_{n}\}$ and a target word\footnote{Referred as a lexical unit when appeared in the Framenet.} $w_t$ in $X$ , the frame identification task aims at predicting the frame $Y_{f}$ evoked by $w_t$ according to the sentence context, while the argument identification task aims at finding the frame-related spans $Y_{s}$ from the context and assigning their semantic roles $Y_{r}$.
In this paper, we introduce frame definitions $k_f$ and role definitions $k_r$ in the FrameNet as knowledge source $K$. 
Frame and role definitions refer to the descriptions of the frame and its corresponding roles, as shown in Figure \ref{fig:intro}. 
Specifically, $k_f$ is used in the frame identification task while $k_r$ is served in the argument identification task.




\section{Methods}


Figure \ref{fig:overall} illustrates the overall architecture of the proposed method. The model comprises a Memory-based Knowledge Extraction Module (MKEM) and a Task-oriented Knowledge Probing Module (TKPM). 
MKEM is responsible for selecting relevant frame knowledge from $K$ and constructing the continuous learnable knowledge vector $P_C$. TKPM accounts for integrating the selected latent knowledge into a PLM and adapting it to frame and argument identification tasks. Details of our methods are described as follows.





\subsection{Memory-based Knowledge Extraction Module}
Since undifferentiated knowledge injection will lead to the knowledge noise problems \cite{liu2020k}, we introduce a simple yet effective end-to-end memory network \cite{sukhbaatar2015end} to extract the most related definition sentences from $K$ as a priori knowledge.

Given knowledge spans $K=\{k_1, k_2, \cdots\}$ to be stored in memory and raw input $X$, MKEM first computes the input memory weight $a_i$ by taking the inner product followed by a softmax, then forms the output memory representation as the continuous prompts $P_C$ using weighted sum, which is:
\begin{eqnarray}
    a_i &=& \text{Softmax}(\text{mean}(e(X))^\top (W_i \cdot \text{mean}(e(k_i)))) \\
    P_C &=& \sum_i a_i (W_o \cdot \text{mean}(e(k_i)))
\end{eqnarray}
where $W_i$ and $W_o$ are learnable parameters, $e(\cdot)$ indicates the shared embedding function, and the token-wise mean function $\text{mean}(\cdot)$ is applied to obtain the sentence vectors. 

Extracting knowledge from the entire $K$ is always time-consuming. In practice, only a relatively relevant subset $K' \in K$ is transferred into the MKEM. For the frame identification task, the definitions of frames associated with $w_t$ are retrieved as $K' \in k_f$. For target words that do not have a corresponding lexical unit in the FrameNet ontology, a simple cosine similarity function with GloVe embeddings is used to extract the most similar lexical units. For the argument identification task, we construct $K' \in k_r$ by fetching the definitions of semantic roles associated with a given frame.

\subsection{Task-oriented Knowledge Probing Module}
The task-oriented knowledge probing module is proposed to leverage the common-sense knowledge from PLMs and inject the frame-related knowledge into the prompting model.

\begin{figure}
    \centering
    \includegraphics[width=0.95\linewidth]{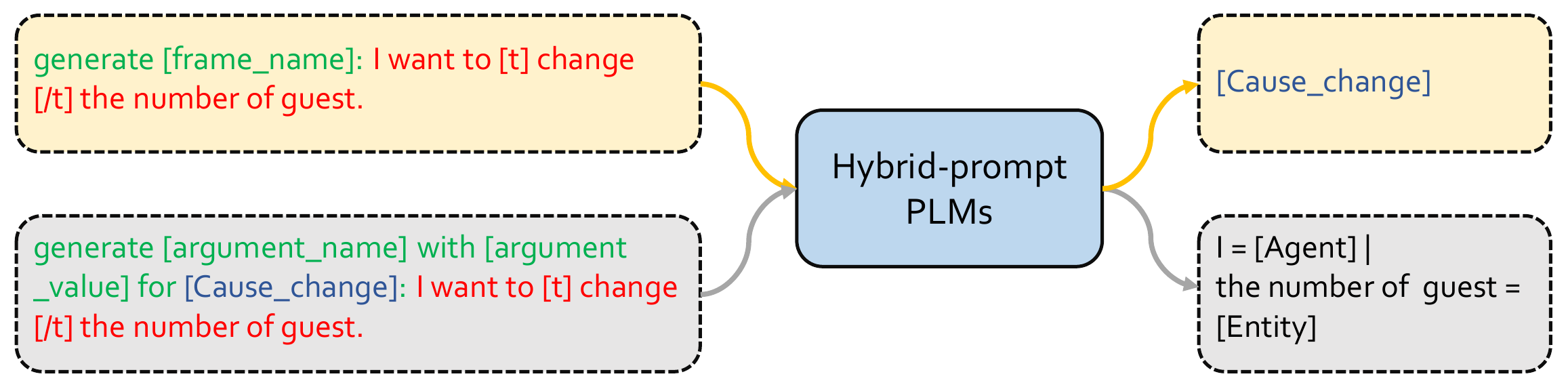}
    \caption{Discrete prompt templates are constructed for two tasks respectively.}
    \label{fig:prompts}
\end{figure}

Specifically, TKPM prepends a prefix for the PLMs to obtain $[P_C;P_D;Y]$, where $P_D=[\text{head}_i; X]$ is the discrete prompt.
Figure \ref{fig:prompts} depicts the prefix template $\text{head}_i$ and its corresponding $Y$ for different tasks: (1) For the frame identification task, $\text{head}_i$ is set to ``\textit{generate [frame\_name]:}" with the frame label $Y_f$ of $X$ as the target output. (2) For the argument identification task, $\text{head}_i$ is set to ``\textit{generate [argument\_name] with [argument\_value] for} $\mathcal{F}$" where $\mathcal{F}$ is the given frame label. Its target output is formed as a sequence of span-role pairs: ``$y_s^1$ = $y_e^1$ $|$ $y_s^2$ = $y_e^2$ $|$ $\cdots$", where $y_s^i \in Y_s$ and $y_r^i \in Y_r$ are frame-related spans and semantic roles, respectively. Such a mechanism can effectively combine knowledge from different sources and allow the MKEM module to be trained while fine-tuning PLM parameters.

\subsection{Model Training}
After constructing the hybrid prompts $P_C$ and $P_D$, we perform prompt-tuning on a PLM in an end-to-end manner. The model is trained with the maximum likelihood objective, defined as:
\begin{equation}
    \mathcal{L}_\theta = - \sum_{i=1}^n \log P_\theta(Y_i | Y_{<i}, X, P_C, P_D)
\end{equation}
where $\theta$ is the model parameters, $X$ represents the raw input, and $Y$ is the target sequence. $P_C$ and $P_D$ indicate continuous and discrete prompts, respectively.

To take full advantage of the prior FrameNet knowledge, we follow \cite{kshirsagar2015frame,zheng2022double} to pre-train the model using exemplar instances. We construct pre-train data for both tasks based on the exemplars, which helps the model obtain a better initial representation for both the frame and role labels. Then, we jointly fine-tune the parameters using original training data.

\section{Experiment}
\label{sec:exp}

\subsection{Datasets and Experimental Settings}
Two benchmark datasets\footnote{https://framenet.icsi.berkeley.edu/fndrupal/about}: FrameNet1.5 and FrameNet1.7 are used for evaluation. FrameNet1.5 defines 1,019 frames and 9,634 frame elements. FrameNet1.7 is an extended version of FrameNet1.5, which contains more fine-grained frames and exemplars, with a total of 1,221 frames and 11,428 frame elements. We follow the same splits of datasets as depicted in open-SESAME \cite{swayamdipta2017frame}. Besides, we follow the way \cite{kshirsagar2015frame,zheng2022double} in pre-training the model: initially using the exemplar instances followed by fine-tuning the model parameters using original training data.


We validate the proposed method based on a T5-base model \cite{raffel2020exploring}. The learning rates for pre-training and prompt-tunning are set to $5\times10^{-4}$ and $2\times10^{-4}$, respectively. We pre-train the model using exemplars for one epoch, followed by prompt-tuning for three epochs using training data. 


The baselines fall into two categories: (1) those don't use frame knowledge \cite{das2014frame,kalyanpur2020open,swayamdipta2017frame,bastianelli2020encoding,marcheggiani2020graph,yang2017joint}; (2) those utilize frame knowledge \cite{jiang2021exploiting,su2021knowledge,zheng2022double}. \cite{zheng2022double} introduce the frame-frame knowledge and design a double-graph based frames to improve the performances.

\subsection{Empirical Results}

\begin{table*}[ht]
\resizebox{\textwidth}{!}{%
    
    \begin{tabular}{| l | c | c | c | c | c | c | c | c | c | c |}
        \hline
        \multirow{4}{*}{\textbf{Model}} & \multicolumn{5}{c |}{\textbf{FrameNet1.5}} & \multicolumn{5}{c |}{\textbf{FrameNet1.7}} \\
        \cline{2-11}
        & \multicolumn{2}{c |}{\textbf{Frame Id}} & \multicolumn{3}{c |}{\textbf{Arg Id}} & \multicolumn{2}{c |}{\textbf{Frame Id}} & \multicolumn{3}{c |}{\textbf{Arg Id}} \\
        \cline{2-11}
       & \multicolumn{2}{c |}{Accuracy} & \multirow{2}{*}{Precision} & \multirow{2}{*}{Recall} & \multirow{2}{*}{F1} & \multicolumn{2}{c |}{Accuracy} & \multirow{2}{*}{Precision} & \multirow{2}{*}{Recall} & \multirow{2}{*}{F1}\\
       \cline{2-3} \cline{7-8}
       & All & Amb.& & & & All & Amb. & & & \\
      \hline
      SEMAFOR \cite{das2014frame} (2014) & 83.6 & 69.2 & 65.6 & 53.8 & 59.1 & - & - & - & - & - \\
      open-SESAME \cite{swayamdipta2017frame} (2017) & 86.9 &  -  & 69.4 & 60.5 & 64.6 & 86.6 &  -  & 60.6 & 58.9 & 59.8 \\
      Yang and Mitchell \cite{yang2017joint} (2017) & 88.2 & 75.7 & 70.2 & 60.2 & 65.5 & - & - & - & - & - \\
      Bastianelli et al.\cite{bastianelli2020encoding} (2020) & 90.1 & - & 74.6 & 74.4 & 74.5 & - & - & - & - & - \\
      Marchegginiani and Titov \cite{marcheggiani2020graph} (2020) & - & - & 69.8 & 68.8 & 69.3 & - & - & - & - & - \\
      Transformer (Full-Gen) \cite{kalyanpur2020open} (2020) & - & - & - & - & - & 87.0 &  -  & 71 & 73 & 72\\
      Transformer (Multi-task) \cite{kalyanpur2020open} (2020) & - & - & - & - & - & 87.5 &  -  & 77 & 77 & 77\\
      FIDO \cite{jiang2021exploiting} (2021) & 91.3 & 81.0 & - & - & - & 92.1 & 83.8 & - & - & - \\
      Su et al. \cite{su2021knowledge} (2021) & 92.1 & 82.3 & - & - & - & 92.4 & 84.4 & - & - & - \\
      KID \cite{zheng2022double} (2022) & 91.7 &  -  & 71.7 & \textbf{79.0} & 75.2 & 91.7 &  -  & 74.1 & 77.3 & 75.6 \\
      \hline
      KAF-SPA & \textbf{92.4} & \textbf{86.6} & \textbf{78.9} & 77.9 & \textbf{78.4} & \textbf{93.6} & \textbf{89.1} & \textbf{81.9} & \textbf{80.7} & \textbf{81.3}\\
      \hline
    \end{tabular}}
    \caption{ Experiments results on the FrameNet1.5 and FrameNet1.7 datasets. ``-'' indicates that the baselines do not report the result. ``All'' indicates a test on the whole test data. ``Amb.'' describes a test on ambiguous test data with two or more candidate frames for a target word. }
    \label{tab:table_fn15}

\end{table*}

\begin{table}[ht]
\centering
\resizebox{0.85\linewidth}{!}{%
\begin{tabular}{|l|clccc|}
\hline
\multicolumn{1}{|c|}{\multirow{4}{*}{\textbf{Model}}} & \multicolumn{5}{c|}{\textbf{FrameNet1.5}}                                                                                                                                                 \\ \cline{2-6} 
\multicolumn{1}{|c|}{}                       & \multicolumn{2}{c|}{\textbf{Frame Id}} & \multicolumn{3}{c|}{\textbf{Arg Id}}                                                                                            \\ \cline{2-6} 
\multicolumn{1}{|c|}{}                       & \multicolumn{2}{c|}{Accuracy}                           & \multicolumn{1}{c|}{\multirow{2}{*}{Precision}} & \multicolumn{1}{c|}{\multirow{2}{*}{Recall}} & \multirow{2}{*}{F1}   \\ \cline{2-3}
\multicolumn{1}{|c|}{}                       & \multicolumn{1}{c|}{All}   & \multicolumn{1}{c|}{Amb.}  & \multicolumn{1}{c|}{}                           & \multicolumn{1}{c|}{}                        &                       \\ \hline
KAF-SPA                                         & \multicolumn{1}{c|}{92.4}  & \multicolumn{1}{c|}{86.6}  & \multicolumn{1}{c|}{78.9}                       & \multicolumn{1}{c|}{77.9}                    & 78.4                  \\
w/o MKEM                                         & \multicolumn{1}{c|}{90.6}  & \multicolumn{1}{c|}{83.4}      & \multicolumn{1}{c|}{77.4}                       & \multicolumn{1}{c|}{77.0}                    & 77.2                  \\
w/o TKPM                                         & \multicolumn{1}{c|}{92.4}      & \multicolumn{1}{c|}{86.0}      & \multicolumn{1}{c|}{77.5}                           & \multicolumn{1}{c|}{76.2}                        &           76.9            \\
w/o PT                                         & \multicolumn{1}{c|}{91.0}  & \multicolumn{1}{c|}{78.7}      & \multicolumn{1}{c|}{77.5}                       & \multicolumn{1}{c|}{75.5}                    & 76.5                  \\ \hline
\multirow{4}{*}{}                            & \multicolumn{5}{c|}{\textbf{FrameNet1.7}}                                                                                                                                                 \\ \cline{2-6} 
                                             & \multicolumn{2}{c|}{\textbf{Frame Id}} & \multicolumn{3}{c|}{\textbf{Arg Id}}                                                                                            \\ \cline{2-6} 
                                             & \multicolumn{2}{c|}{Accuracy}                           & \multicolumn{1}{l|}{\multirow{2}{*}{Precision}} & \multicolumn{1}{c|}{\multirow{2}{*}{Recall}} & \multirow{2}{*}{F1}   \\ \cline{2-3}
                                             & \multicolumn{1}{c|}{All}   & \multicolumn{1}{c|}{Amb.}  & \multicolumn{1}{l|}{}                           & \multicolumn{1}{c|}{}                        &                       \\ \hline
KAF-SPA                                         & \multicolumn{1}{c|}{93.6}  & \multicolumn{1}{c|}{89.1}  & \multicolumn{1}{c|}{81.9}                       & \multicolumn{1}{c|}{80.7}                    & 81.3                  \\
w/o MKEM                                         & \multicolumn{1}{c|}{90.9}  & \multicolumn{1}{c|}{86.1}      & \multicolumn{1}{c|}{80.6}                       & \multicolumn{1}{c|}{80.7}                    & 80.7                  \\
w/o TKPM                    & \multicolumn{1}{c|}{93.5}      & \multicolumn{1}{c|}{88.7}      & \multicolumn{1}{c|}{80.0}                           & \multicolumn{1}{c|}{76.9}                        & \multicolumn{1}{c|}{78.5} \\
w/o PT                                         & \multicolumn{1}{c|}{91.2}  & \multicolumn{1}{c|}{77.6}      & \multicolumn{1}{c|}{76.0}                       & \multicolumn{1}{c|}{74.3}                    & 75.1                  \\ \hline

\end{tabular}}%
\caption{The results of the ablation study on model components. ``w/o MKEM'' indicates the study removing the memory-based knowledge extraction module. ``w/o TKPM'' illustrates a test leaving discrete prompts out in the task-oriented knowledge probing module and training two tasks separately. ``w/o PT'' indicates a study refraining using the exemplar-based pre-train process.}
\label{tab:ablation}
\end{table}
For the frame identification (Frame Id) task, we apply the \textbf{accuracy} to calculate the ratio of the correctly predicted frame on both total and ambiguous (Amb.) test data.
For the argument identification (Arg Id) task, we use \textbf{Precision}, \textbf{Recall} and \textbf{F1} to evaluate the quality of extracted spans and the according frame element roles. 
Table \ref{tab:table_fn15} shows the performance of the proposed method compared to the results from different baselines. 
From Table \ref{tab:table_fn15}, FIDO, \cite{su2021knowledge} and KID models perform considerably better than other baselines without the knowledge. 
Incorporating knowledge contributes both to the frame and argument identification performance consistently.
Compared with the knowledge-augmented strong baselines, our method achieves state-of-art in the frame identification task. 
The main result table (Table \ref{tab:table_fn15}) also shows that our method outperforms the best baseline by a wide margin in identifying ambiguous frames (86.6 vs. 82.3 for FrameNet1.5 and 89.1 vs. 84.4 for FrameNet1.7). 
Besides, in the arguments identification task, the proposed method significantly outperforms the best baselines: 3$\%$ in F1 of FrameNet1.5 and over 5$\%$ in the FrameNet1.7 dataset. The results confirm the capability of KAF-SPA in selecting/incorporating accurate knowledge and probing task-relevant knowledge in PLMs to improve frame semantic parsing performance. Another observation is that our method shows an inferior recall performance in FrameNet1.5, potentially due to the imbalanced class distributions in Framenet1.5. KID has a high recall performance but a low precision score, indicating that the model lacks the capability to differentiate false positive outcomes. 

\subsection{Ablation Study}

    
        



We conduct an ablation test on the two datasets to investigate the influence of the proposed modules and the exemplar-based pre-train process. Table \ref{tab:ablation} demonstrates the ablation results.

\textbf{w/o Memory-based Knowledge Extraction Module} 
In Table \ref{tab:ablation}, the performance drops over 1$\%$, indicating that selecting proper knowledge structures benefits the frame semantic parsing. Furthermore, the model performs better in all metrics than the baselines under the no-knowledge setting.

\textbf{w/o Task-oriented Knowledge Probing Module} 
Compared to the whole model, the ``w/o TKPM'' has a slight performance degradation, demonstrating that joint learning of the two subtasks helps improve the model's overall performance.
Moreover, we also verify that the exemplar-based pre-train method is beneficial for improving frame semantic parsing.

\section{Conclusion}
\label{sec:conclusion}
This paper proposes KAF-SPA to enhance semantic representation and disambiguate semantic frames. Experimental results on two public FrameNet datasets demonstrate that our method outperforms strong baselines by a wide margin, achieve state-of-art results. In future work, we will explore effective ways to incorporate semantic information into pre-training for natural language understanding and reasoning.

\vfill\pagebreak




\begin{thebibliography}{10}

\bibitem{baker1998berkeley}
Collin~F Baker, Charles~J Fillmore, and John~B Lowe,
\newblock ``The berkeley framenet project,''
\newblock in {\em COLING 1998 Volume 1: The 17th International Conference on
  Computational Linguistics}, 1998.

\bibitem{das2014frame}
Dipanjan Das, Desai Chen, Andr{\'e}~FT Martins, Nathan Schneider, and Noah~A
  Smith,
\newblock ``Frame-semantic parsing,''
\newblock {\em Computational linguistics}, vol. 40, no. 1, pp. 9--56, 2014.

\bibitem{chen2013unsupervised}
Yun-Nung Chen, William~Yang Wang, and Alexander~I Rudnicky,
\newblock ``Unsupervised induction and filling of semantic slots for spoken
  dialogue systems using frame-semantic parsing,''
\newblock in {\em 2013 IEEE Workshop on Automatic Speech Recognition and
  Understanding}. IEEE, 2013, pp. 120--125.

\bibitem{guo2020frame}
Shaoru Guo, Ru~Li, Hongye Tan, Xiaoli Li, Yong Guan, Hongyan Zhao, and Yueping
  Zhang,
\newblock ``A frame-based sentence representation for machine reading
  comprehension,''
\newblock in {\em Proceedings of the 58th Annual Meeting of the Association for
  Computational Linguistics}, 2020, pp. 891--896.

\bibitem{shen2007using}
Dan Shen and Mirella Lapata,
\newblock ``Using semantic roles to improve question answering,''
\newblock in {\em Proceedings of the 2007 joint conference on empirical methods
  in natural language processing and computational natural language learning
  (EMNLP-CoNLL)}, 2007, pp. 12--21.

\bibitem{kalyanpur2020open}
Aditya Kalyanpur, Or~Biran, Tom Breloff, Jennifer Chu-Carroll, Ariel Diertani,
  Owen Rambow, and Mark Sammons,
\newblock ``Open-domain frame semantic parsing using transformers,''
\newblock {\em arXiv preprint arXiv:2010.10998}, 2020.

\bibitem{petroni2019language}
Fabio Petroni, Tim Rockt{\"a}schel, Sebastian Riedel, Patrick Lewis, Anton
  Bakhtin, Yuxiang Wu, and Alexander Miller,
\newblock ``Language models as knowledge bases?,''
\newblock in {\em EMNLP-IJCNLP}, 2019, pp. 2463--2473.

\bibitem{li2021prefix}
Xiang~Lisa Li and Percy Liang,
\newblock ``Prefix-tuning: Optimizing continuous prompts for generation,''
\newblock in {\em Proceedings of the 59th Annual Meeting of the Association for
  Computational Linguistics and the 11th International Joint Conference on
  Natural Language Processing (Volume 1: Long Papers)}, 2021, pp. 4582--4597.

\bibitem{kshirsagar2015frame}
Meghana Kshirsagar, Sam Thomson, Nathan Schneider, Jaime~G Carbonell, Noah~A
  Smith, and Chris Dyer,
\newblock ``Frame-semantic role labeling with heterogeneous annotations,''
\newblock in {\em Proceedings of the 53rd Annual Meeting of the Association for
  Computational Linguistics and the 7th International Joint Conference on
  Natural Language Processing (Volume 2: Short Papers)}, 2015, pp. 218--224.

\bibitem{swayamdipta2017frame}
Swabha Swayamdipta, Sam Thomson, Chris Dyer, and Noah~A Smith,
\newblock ``Frame-semantic parsing with softmax-margin segmental rnns and a
  syntactic scaffold,''
\newblock {\em arXiv preprint arXiv:1706.09528}, 2017.

\bibitem{bastianelli2020encoding}
Emanuele Bastianelli, Andrea Vanzo, and Oliver Lemon,
\newblock ``Encoding syntactic constituency paths for frame-semantic parsing
  with graph convolutional networks,''
\newblock {\em arXiv preprint arXiv:2011.13210}, 2020.

\bibitem{marcheggiani2020graph}
Diego Marcheggiani and Ivan Titov,
\newblock ``Graph convolutions over constituent trees for syntax-aware semantic
  role labeling,''
\newblock in {\em Proceedings of the 2020 Conference on Empirical Methods in
  Natural Language Processing (EMNLP)}, 2020, pp. 3915--3928.

\bibitem{jiang2021exploiting}
Tianyu Jiang and Ellen Riloff,
\newblock ``Exploiting definitions for frame identification,''
\newblock in {\em Proceedings of the 16th Conference of the European Chapter of
  the Association for Computational Linguistics: Main Volume}, 2021, pp.
  2429--2434.

\bibitem{su2021knowledge}
Xuefeng Su, Ru~Li, Xiaoli Li, Jeff~Z Pan, Hu~Zhang, Qinghua Chai, and Xiaoqi
  Han,
\newblock ``A knowledge-guided framework for frame identification,''
\newblock in {\em ACL-IJCNLP (Volume 1: Long Papers)}, 2021, pp. 5230--5240.

\bibitem{zheng2022double}
Ce~Zheng, Xudong Chen, Runxin Xu, and Baobao Chang,
\newblock ``A double-graph based framework for frame semantic parsing,''
\newblock in {\em NAACL}, 2022, pp. 4998--5011.

\bibitem{liu2020k}
Weijie Liu, Peng Zhou, Zhe Zhao, Zhiruo Wang, Qi~Ju, Haotang Deng, and Ping
  Wang,
\newblock ``K-bert: Enabling language representation with knowledge graph,''
\newblock in {\em Proceedings of the AAAI Conference on Artificial
  Intelligence}, 2020, vol.~34, pp. 2901--2908.

\bibitem{sukhbaatar2015end}
Sainbayar Sukhbaatar, Jason Weston, Rob Fergus, et~al.,
\newblock ``End-to-end memory networks,''
\newblock {\em Advances in neural information processing systems}, vol. 28,
  2015.

\bibitem{raffel2020exploring}
Colin Raffel, Noam Shazeer, Adam Roberts, Katherine Lee, Sharan Narang, Michael
  Matena, Yanqi Zhou, Wei Li, and Peter~J Liu,
\newblock ``Exploring the limits of transfer learning with a unified
  text-to-text transformer,''
\newblock {\em Journal of Machine Learning Research}, vol. 21, pp. 1--67, 2020.

\bibitem{yang2017joint}
Bishan Yang and Tom Mitchell,
\newblock ``A joint sequential and relational model for frame-semantic
  parsing,''
\newblock in {\em Proceedings of the 2017 Conference on Empirical Methods in
  Natural Language Processing}, 2017, pp. 1247--1256.

\end{thebibliography}

\end{document}